\documentclass[11pt]{article}
\usepackage[preprint]{neurips_2025}
\usepackage{tikz}
\usetikzlibrary{arrows.meta,positioning,shapes.geometric,fit,calc}
\usepackage{amsmath}
\usepackage{amsfonts}
\usepackage{booktabs}
\usepackage{float}
\usepackage[colorlinks=true, urlcolor=blue, linkcolor=blue, citecolor=blue]{hyperref}

\makeatletter
\def\@trackname{}
\makeatother

\title{Non-Interfering Weight Fields: Treating Model Parameters as a Continuously Extensible Function}

\author{
Sarim Chaudhry \\
Purdue University \\
\texttt{chaud158@purdue.edu}
}

\date{}

\begin{document}

\maketitle

\begin{abstract}
Large language models store all learned knowledge in a single, fixed weight vector. Teaching a model new capabilities requires modifying those same weights, inevitably degrading previously acquired knowledge. This fundamental limitation, known as catastrophic forgetting, has resisted principled solutions for decades. Existing approaches treat weights as immutable artifacts that must be protected through techniques like regularization heuristics, replay buffers, or isolated adapter modules. The problem is none of these provide a structural guarantee against forgetting. In this work, we propose Non-Interfering Weight Fields (NIWF), a framework that replaces the fixed weight paradigm with a learned function that generates weight configurations on demand from a continuous capability coordinate space. After training on a task, we commit the occupied coordinate region by snapshotting the field's outputs on anchor points to enforce a functional lock during all future training. We validate NIWF on sequential instruction-following and code generation tasks using Mistral-7B, demonstrating zero forgetting on committed tasks with competitive perplexity on new tasks. The framework introduces the notion of software-like versioning for neural network intelligence, where capabilities can be committed, extended, composed, and rolled back without retraining.
\end{abstract}

\section{Introduction}

\begin{figure*}[t]
\centering
\includegraphics[width=\textwidth]{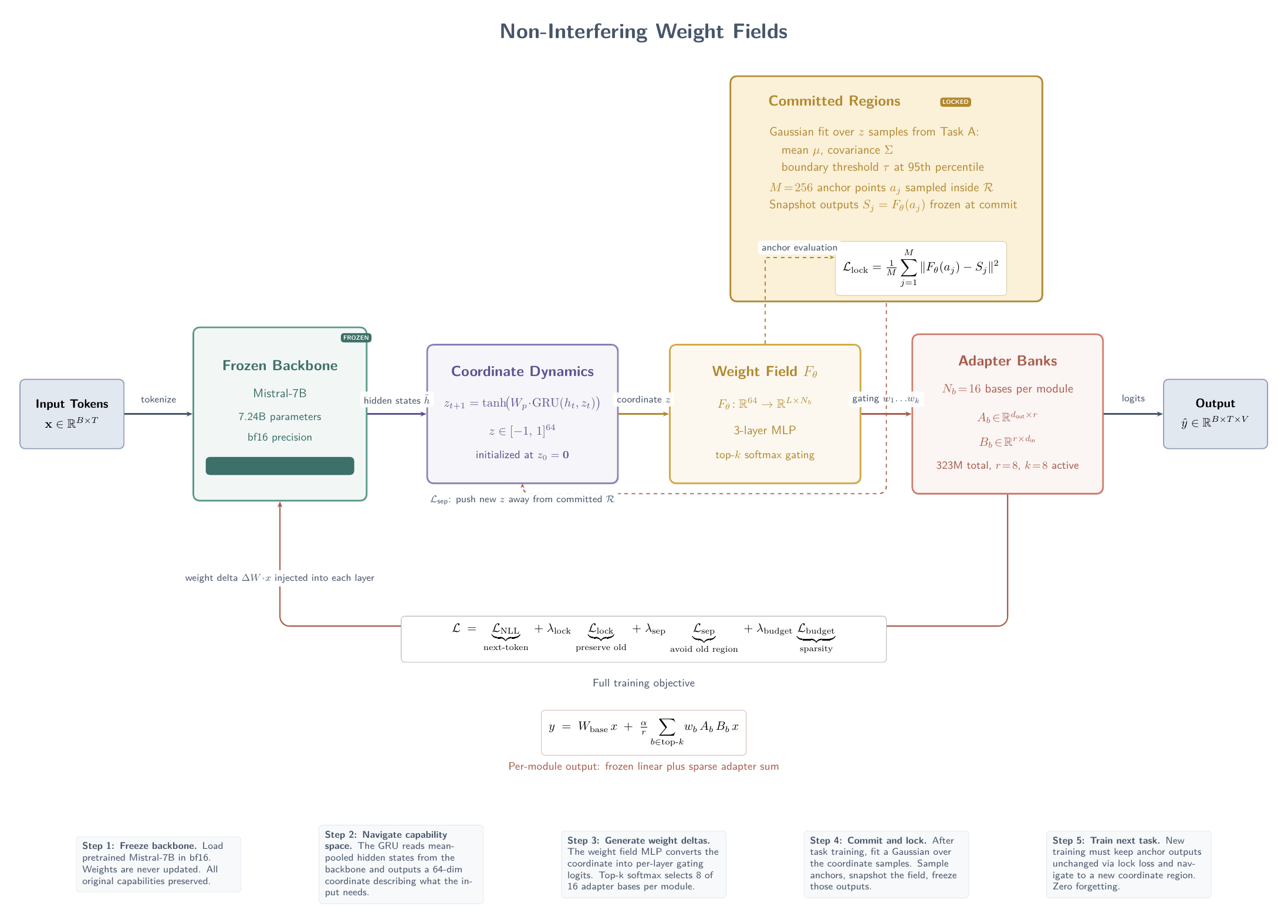}
\caption{The NIWF architecture. A frozen backbone processes tokens
through a two-pass forward. The coordinate dynamics module produces
a capability coordinate $z$ from mean-pooled hidden states. The
weight field maps $z$ to sparse gating over low-rank adapter bases.
After training, coordinate regions are committed and functionally
locked via anchor-based snapshot constraints.}
\label{fig:architecture}
\end{figure*}

Consider the current workflow for improving a deployed language model. A team identifies a deficiency in the model's mathematical reasoning. They collect training data, fine-tune the model, and evaluate. The math performance will improve, but now it has subtly become worse at creative writing, code generation, and a dozen other capabilities. This is the default behavior of gradient-based learning on neural networks, and it has been documented extensively since the earliest connectionist systems \citep{mccloskey1989catastrophic, ratcliff1990connectionist, french1999catastrophic}.

A neural network encodes everything it knows in a single weight vector $W \in \mathbb{R}^d$. Every gradient update to teach a new capability modifies the same numbers that encode existing capabilities. There is no separation between the substrate and the content. This results in every write being potentially destructive.

The field has developed numerous techniques to mitigate this problem. Elastic Weight Consolidation (EWC) \citep{kirkpatrick2017overcoming} penalizes changes to weights deemed important for prior tasks, but these importance estimates can be noisy and the penalty is often too soft. Progressive Neural Networks \citep{rusu2016progressive} freeze old columns and add new ones, but parameter cost grows linearly with the number of tasks. Experience replay \citep{rolnick2019experience, chaudhry2019continual} maintains a buffer of old data, but provides no formal guarantee against forgetting. Low-rank adapters (LoRA) \citep{hu2021lora} and their variants \citep{dettmers2023qlora, zhang2023adalora} offer parameter-efficient fine-tuning but are static patches with no routing, no versioning, and no continual learning story. Mixture-of-Experts (MoE) architectures \citep{shazeer2017outrageously, fedus2022switch, dai2024deepseekmoe} route tokens to discrete experts but have no concept of committing regions or preventing interference across training stages.

None of these approaches address the fundamental issue. They all treat the weight vector as the primary object and try to protect it through external mechanisms. We propose a different starting point entirely.

In this paper, we introduce Non-Interfering Weight Fields (NIWF), a framework that replaces the fixed weight paradigm with a learned, continuously extensible function over a capability coordinate space. The core insight is simple but consequential. Instead of storing knowledge as weights, we store it as a function that generates weights. Different inputs navigate to different coordinates in this space, and different coordinates produce different weight configurations. New learning extends the function into unoccupied territory rather than overwriting existing mappings. Old regions are functionally frozen, making catastrophic forgetting structurally impossible rather than merely unlikely.

Our contributions are as follows.

\begin{itemize}
\item We formalize the concept of a weight field, a learned mapping $F_\theta: \mathbb{R}^{d_z} \to \mathbb{R}^{L \times N_b}$ that generates sparse gating decisions over banks of low-rank adapter bases conditioned on a capability coordinate $z$. We show that this formulation cleanly separates stored capacity (total bases) from active computation (top-$k$ selection), enabling monotonic capacity growth without proportional compute increase.

\item We introduce a region commitment protocol that provides a hard, non-approximate guarantee against catastrophic forgetting. After training on a task, the occupied region of coordinate space is snapshotted and functionally locked. During all future training, the field must reproduce the exact same outputs for coordinates within the locked region. This is enforced through a mean-squared-error loss on anchor points.

\item We design a recurrent coordinate dynamics module that navigates the capability space conditioned on the model's hidden states, enabling input-dependent weight generation without external task labels or routing signals.

\item We implement and validate the complete framework on Mistral-7B \citep{jiang2023mistral}, demonstrating the complete process of train, commit, extend, and evaluate on sequential instruction-following and code generation tasks with zero degradation on committed capabilities.
\end{itemize}

\section{Related Work}

\subsection{Catastrophic Forgetting and Continual Learning}

The problem of catastrophic forgetting was identified in early connectionist research \citep{mccloskey1989catastrophic, ratcliff1990connectionist} and has remained central to neural network research ever since. \citet{french1999catastrophic} provides a comprehensive survey of early approaches. Modern continual learning methods broadly fall into three categories.

Regularization-based methods add terms to the loss function that penalize changes to parameters important for prior tasks. EWC \citep{kirkpatrick2017overcoming} uses the diagonal of the Fisher information matrix as an importance measure. Synaptic Intelligence (SI) \citep{zenke2017continual} accumulates per-parameter importance online during training. Memory Aware Synapses (MAS) \citep{aljundi2018memory} estimates importance based on sensitivity of learned function outputs. These methods provide soft protection that degrades as the number of tasks grows, since the regularization terms from different tasks eventually conflict.

Replay-based methods maintain a buffer of examples from prior tasks and interleave them during training on new tasks. Experience replay \citep{rolnick2019experience}, A-GEM \citep{chaudhry2019continual}, and DER++ \citep{buzzega2020dark} represent this family. While effective in practice, replay does not provide guarantees and requires storing potentially sensitive data from prior tasks.

Architecture-based methods allocate dedicated capacity for each task. Progressive Neural Networks \citep{rusu2016progressive} add new columns for each task with lateral connections. PackNet \citep{mallya2018packnet} prunes and freezes subnetworks. These methods provide stronger guarantees but scale poorly. NIWF is closest in spirit to architecture-based methods but achieves the separation through a continuous functional mapping rather than discrete module allocation, avoiding the linear parameter growth problem.

\subsection{Parameter-Efficient Fine-Tuning}

LoRA \citep{hu2021lora} introduced the idea of freezing pretrained weights and training low-rank additive updates. Subsequent work has explored quantized variants \citep{dettmers2023qlora}, adaptive rank allocation \citep{zhang2023adalora}, and composition of multiple adapters \citep{huang2023lorahub}. However, all LoRA variants produce a single static delta that is applied identically regardless of input. There is no mechanism for input-dependent routing, no notion of region locking, and no built-in continual learning capability. NIWF can be viewed as a conditional, routed, and versionable generalization of LoRA, where the static delta is replaced by a dynamically generated mixture over a bank of low-rank bases.

\subsection{Mixture-of-Experts}

Sparse MoE models \citep{shazeer2017outrageously, lepikhin2020gshard, fedus2022switch, dai2024deepseekmoe} route tokens to specialized expert subnetworks via a learned gating function. Recent work has explored gating mechanisms in attention specifically, including Switch Heads \citep{csordas2024switchhead}, Native Sparse Attention \citep{yuan2025native}, and gated attention variants \citep{qiu2025gated}. These architectures demonstrate the value of conditional computation but operate at the token level within a single training stage. They have no concept of committing routing decisions across training stages, and expert parameters remain mutable throughout training.

\subsection{Hypernetworks}

Hypernetworks \citep{ha2016hypernetworks} generate weights for a target network using a secondary network. HyperTransformers \citep{ivison2022hyperdecoders} and related work \citep{von2019continual} have explored hypernetworks for few-shot learning and task-conditional generation. NIWF extends this line of work by (a) generating sparse gating decisions over low-rank bases rather than full weight matrices, making the approach practical for large language models, (b) introducing a continuous coordinate space with recurrent dynamics, and (c) adding region commitment and functional locking as first-class operations.

\section{Method}

We now describe the NIWF framework in detail. The system consists of four components, a frozen pretrained backbone $\mathcal{B}$, a bank of low-rank adapter bases $\{A_b, B_b\}_{b=1}^{N_b}$ per target module, a weight field $F_\theta$ that maps capability coordinates to gating decisions, and a coordinate dynamics module $G_\phi$ that navigates the coordinate space conditioned on the model's internal representations.

\subsection{Frozen Backbone with Conditional Adapter Banks}

Let $\mathcal{B}$ denote a pretrained language model with all parameters frozen. We identify a set of target linear modules within $\mathcal{B}$, typically the query, key, value, output, gate, up, and down projections in each transformer layer. For a target module with weight matrix $W_{\text{base}} \in \mathbb{R}^{d_{\text{out}} \times d_{\text{in}}}$, we introduce a bank of $N_b$ low-rank adapter bases, each parameterized by a pair of factor matrices:
\begin{equation}
A_b \in \mathbb{R}^{d_{\text{out}} \times r}, \quad B_b \in \mathbb{R}^{r \times d_{\text{in}}}, \quad b = 1, \ldots, N_b
\end{equation}
where $r \ll \min(d_{\text{in}}, d_{\text{out}})$ is the rank. Each pair defines a low-rank weight delta $\Delta W_b = A_b B_b \in \mathbb{R}^{d_{\text{out}} \times d_{\text{in}}}$. Following \citet{hu2021lora}, we initialize $B_b$ with Kaiming uniform and $A_b$ with zeros, ensuring that the initial delta is zero and training starts from the pretrained model's behavior.

Given a gating vector that selects the top-$k$ bases with indices $\mathcal{S} \subset \{1, \ldots, N_b\}$ and associated weights $\{w_b\}_{b \in \mathcal{S}}$ (normalized via softmax over the selected logits), the effective output of the adapted module on input $x \in \mathbb{R}^{B \times T \times d_{\text{in}}}$ is:
\begin{equation}
y = W_{\text{base}} x + \frac{\alpha}{r} \sum_{b \in \mathcal{S}} w_b \cdot x B_b^\top A_b^\top
\label{eq:niwf_forward}
\end{equation}
where $\alpha$ is a scaling hyperparameter analogous to the LoRA scaling factor.

A key implementation detail concerns memory efficiency. A naive implementation would gather the selected adapter matrices as tensors of shape $[B, T, k, d_{\text{out}}, r]$, which for MLP projections with $d_{\text{out}} = 14336$ would consume nearly 1 GB per module call. We avoid this by performing sequence-level gating, where the same set of bases is selected for all tokens in a sequence. The gating indices have shape $[B, k]$ rather than $[B, T, k]$, and the gathered matrices have shape $[B, k, d_{\text{out}}, r]$, reducing memory by a factor of $T$. The factored computation then proceeds as:
\begin{align}
h &= \text{einsum}(\texttt{`bti,bkri->btkr'}, x, B_{\mathcal{S}}) \label{eq:step1} \\
\delta &= \text{einsum}(\texttt{`btkr,bkor->btko'}, h, A_{\mathcal{S}}) \label{eq:step2} \\
y &= W_{\text{base}} x + \frac{\alpha}{r} \sum_{j=1}^{k} w_j \cdot \delta_{:,:,j,:} \label{eq:step3}
\end{align}
where $A_{\mathcal{S}}, B_{\mathcal{S}}$ denote the gathered subsets of adapter factor matrices.

\subsection{Weight Field}

The weight field $F_\theta: \mathbb{R}^{d_z} \to \mathbb{R}^{L \times N_b}$ is a small neural network that maps a capability coordinate $z \in \mathbb{R}^{d_z}$ to gating logits over the adapter bases for each of the $L$ transformer layers. The architecture consists of a three-layer MLP with LayerNorm at the input and GELU activations:
\begin{equation}
F_\theta(z) = W_3 \cdot \text{GELU}(W_2 \cdot \text{GELU}(W_1 \cdot \text{LN}(z)))
\end{equation}
where $W_1 \in \mathbb{R}^{d_h \times d_z}$, $W_2 \in \mathbb{R}^{d_h \times d_h}$, $W_3 \in \mathbb{R}^{(L \cdot N_b) \times d_h}$, and $d_h$ is a hidden dimension (256 in our experiments). The output is reshaped to $[B, L, N_b]$, providing per-layer gating logits.

For each layer $\ell$, the top-$k$ bases are selected and their weights are computed via softmax over the selected logits:
\begin{equation}
\mathcal{S}_\ell = \text{top-}k(F_\theta(z)_\ell), \quad w_b = \frac{\exp(F_\theta(z)_{\ell,b})}{\sum_{b' \in \mathcal{S}_\ell} \exp(F_\theta(z)_{\ell,b'})}, \quad b \in \mathcal{S}_\ell
\end{equation}

This design cleanly separates stored capacity from active computation. The total number of adapter bases $N_b$ determines how much conditional knowledge the system can hold, while $k$ determines how much computation is spent per forward pass. Increasing $N_b$ adds capacity without increasing per-token cost.

\subsection{Coordinate Dynamics}

The capability coordinate $z$ must be set appropriately for each input. Rather than requiring external task labels or manual routing, we learn a recurrent dynamics module $G_\phi$ that evolves $z$ based on the model's own hidden state representations:
\begin{equation}
z_{t+1} = \tanh(W_p \cdot \text{LN}(\text{GRU}(h_t, z_t)))
\end{equation}
where $h_t$ is a summary of the backbone's hidden states (mean-pooled over the sequence dimension from the final layer), GRU is a Gated Recurrent Unit \citep{cho2014learning}, LN is LayerNorm, and $W_p \in \mathbb{R}^{d_z \times d_z}$ is a learned projection. The $\tanh$ nonlinearity bounds $z$ to $[-1, 1]^{d_z}$, preventing coordinate drift.

In our implementation, the coordinate update operates at the sequence level using a two-pass strategy. On the first forward pass, the backbone processes the input with the initial coordinate $z_0 = \mathbf{0}$, producing hidden states that are used to update $z_0 \to z_1$. A second forward pass with $z_1$ produces the final output. This two-pass approach avoids the complexity of per-token coordinate updates within the backbone's forward pass. The hidden states used for the coordinate update are detached from the computation graph to prevent gradients from flowing back through the backbone.

\subsection{Training Objective}

The training loss combines four terms:
\begin{equation}
\mathcal{L} = \mathcal{L}_{\text{NLL}} + \lambda_{\text{lock}} \cdot \mathcal{L}_{\text{lock}} + \lambda_{\text{sep}} \cdot \mathcal{L}_{\text{sep}} + \lambda_{\text{budget}} \cdot \mathcal{L}_{\text{budget}}
\end{equation}

The next-token prediction loss $\mathcal{L}_{\text{NLL}}$ is the standard cross-entropy loss over the vocabulary. This is the primary learning signal that shapes the adapter bases and trains the weight field and coordinate dynamics.

The lock loss $\mathcal{L}_{\text{lock}}$ enforces functional preservation of committed regions (detailed in Section~\ref{sec:commit}). When no regions have been committed (e.g., during the first task), $\lambda_{\text{lock}} = 0$.

The separation loss $\mathcal{L}_{\text{sep}}$ encourages the current task's coordinates to occupy regions distinct from previously committed regions:
\begin{equation}
\mathcal{L}_{\text{sep}} = \mathbb{E}_{z \sim p_{\text{current}}} \left[ \max\left(0, \tau_A + m - d_M(z, \mu_A, \Sigma_A)\right)^2 \right]
\end{equation}
where $d_M$ is the Mahalanobis distance to the committed region with mean $\mu_A$ and covariance $\Sigma_A$, $\tau_A$ is the region threshold, and $m \geq 0$ is an optional margin. This quadratic penalty activates only when coordinates fall within or too close to the committed region, pushing them outward without constraining where they end up.

The budget loss $\mathcal{L}_{\text{budget}} = k$ is currently a constant since top-$k$ is fixed, but it allows for future extensions with dynamic, input-dependent sparsity.

\subsection{Region Commitment and Functional Locking}
\label{sec:commit}

The commitment protocol is the mechanism that converts NIWF from a conditional adapter system into a continual learning framework with provable non-interference. After training on a task, the following procedure is executed.

First, all training examples are passed through the model to collect the set of capability coordinates $\{z_i\}_{i=1}^{N}$ produced by the coordinate dynamics module. These coordinates represent the region of capability space that the task occupies.

Second, a Gaussian model is fit to the collected coordinates, yielding a mean $\mu \in \mathbb{R}^{d_z}$ and covariance $\Sigma \in \mathbb{R}^{d_z \times d_z}$. The region boundary is defined using the Mahalanobis distance at the 95th percentile of the training distribution:
\begin{equation}
\mathcal{R} = \left\{ z \in \mathbb{R}^{d_z} : d_M(z, \mu, \Sigma) \leq \tau \right\}, \quad \tau = Q_{0.95}\left(\{d_M(z_i, \mu, \Sigma)\}_{i=1}^{N}\right)
\end{equation}
where $Q_{0.95}$ denotes the 95th percentile.

Third, a set of anchor points $\{a_j\}_{j=1}^{M}$ is sampled from within $\mathcal{R}$, either by sampling from the fitted Gaussian and rejecting points outside the region, or by subsampling the empirical coordinate set. The field's outputs on these anchors are recorded as the snapshot:
\begin{equation}
S_j = F_\theta(a_j), \quad j = 1, \ldots, M
\end{equation}

During all subsequent training stages, the lock loss enforces that the field reproduces the same outputs on the anchors:
\begin{equation}
\mathcal{L}_{\text{lock}} = \frac{1}{M} \sum_{j=1}^{M} \| F_\theta(a_j) - S_j \|^2
\label{eq:lock_loss}
\end{equation}

This is a hard functional constraint. If the lock loss is driven to zero, the field produces exactly the same gating logits for any coordinate within the committed region. Since the same logits produce the same top-$k$ selection and the same softmax weights, and the base parameters $A_b, B_b$ are also constrained by the lock (they cannot change their contribution to locked gating patterns without changing the field output, which is penalized), the effective weight delta for any input that maps to the committed region is preserved exactly.

The key distinction from regularization-based continual learning methods is that the lock loss operates on the gating function's outputs at specific points, not on aggregate parameter importance estimates. This means the guarantee is local and precise. The field is free to change its behavior arbitrarily for coordinates outside committed regions.

\subsection{Sequential Task Learning Protocol}

The full multi-task protocol proceeds as follows. On the first task, training proceeds normally with $\lambda_{\text{lock}} = 0$ and $\lambda_{\text{sep}} = 0$. After convergence, the region is committed. On each subsequent task, training proceeds with the lock loss active for all previously committed regions and the separation loss pushing coordinates away from those regions. After convergence, the new task's region is committed, and the process repeats.

This protocol ensures that the system's knowledge grows monotonically. Each committed region represents a frozen capability that cannot be degraded by future training. The coordinate space acts as an address book for capabilities, and the commitment protocol acts as a write-once lock that prevents overwriting.

\section{Model Architecture }

\subsection{Adapter Bank Parameterization}

Each target module in the transformer backbone is wrapped with an NIWFLinear layer that contains the adapter bank. For a Mistral-7B model, the target modules span all seven linear projections in each of the 32 transformer layers, specifically the query, key, value, and output projections in the attention block and the gate, up, and down projections in the MLP block. This gives a total of $32 \times 7 = 224$ NIWFLinear modules.

Each NIWFLinear module maintains its own independent bank of $N_b$ adapter bases. The bases are not shared across modules, allowing each projection to specialize its conditional capacity independently. The total parameter count for the adapter banks is:
\begin{equation}
P_{\text{adapt}} = L_{\text{layers}} \cdot \sum_{m \in \text{modules}} N_b \cdot r \cdot (d_{\text{in}}^{(m)} + d_{\text{out}}^{(m)})
\end{equation}

For Mistral-7B with $N_b = 16$ and $r = 8$, this evaluates to approximately 1.29 GB in 32-bit floating point. With AdamW optimizer states (momentum and variance), the total adapter-related memory is approximately 3.88 GB.

\subsection{Weight Field Architecture}

The weight field is a three-layer MLP with $d_z = 64$ input dimension, hidden dimension $d_h = 256$, and output dimension $L \times N_b = 32 \times 16 = 512$ (when not sharing across layers). LayerNorm is applied at the input, and GELU activations are used between layers. The total parameter count of the weight field is approximately 200K, which is negligible compared to the adapter banks and backbone.

An important design decision is whether to share gating decisions across layers (one set of logits for all layers) or produce per-layer gating. Per-layer gating allows different transformer blocks to activate different base combinations for the same input, which is more expressive. In our experiments, we use per-layer gating unless otherwise specified.

\subsection{Coordinate Dynamics Architecture}

The coordinate dynamics module consists of a GRUCell with input size $d_{\text{hidden}} = 4096$ (Mistral-7B's hidden dimension) and hidden size $d_z = 64$, followed by LayerNorm, a linear projection, and a $\tanh$ activation. The hidden states from the backbone's final layer are mean-pooled over the sequence dimension to produce the input to the GRU. The GRU's hidden state is the current coordinate $z$.

The $\tanh$ bounding is important for the region commitment protocol. Without it, coordinates could drift to arbitrarily large values during training, making the Gaussian region model a poor fit. With $\tanh$, all coordinates lie within the hypercube $[-1, 1]^{64}$, and the Gaussian model provides a reasonable approximation to the empirical distribution.

\subsection{Two-Pass Forward Strategy}

The NIWF forward pass requires knowing the capability coordinate $z$ before the backbone's forward pass (since $z$ determines the gating decisions), but $z$ depends on the backbone's hidden states (which are produced during the forward pass). We resolve this circular dependency with a two-pass strategy.

In the first pass, the backbone processes the input with the initial coordinate $z_0 = \mathbf{0}$, producing hidden states $h^{(1)}$. These hidden states are mean-pooled and fed to the dynamics module to produce $z_1 = G_\phi(z_0, \bar{h}^{(1)})$.

In the second pass, the backbone processes the same input with gating decisions derived from $z_1$, producing the final output and loss. The hidden states from the first pass are detached from the computation graph to prevent gradients from flowing back through the backbone during the $z$ update step.

This strategy doubles the number of backbone forward passes during training, but since the backbone is frozen and no gradients flow through it, each pass is relatively cheap. At inference time, the same two-pass strategy is used, adding additional latency compared to the base model. Future work could explore amortized or cached strategies to reduce this overhead.

\section{Experiments}

\subsection{Experimental Setup}

We validate NIWF on a two-task sequential learning scenario using Mistral-7B-v0.1 \citep{jiang2023mistral} as the frozen backbone. The two tasks are:

Task A (instruction following): We train on 4,000 examples from the Alpaca dataset \citep{taori2023alpaca}, which contains general-purpose instruction-response pairs covering diverse topics including summarization, question answering, creative writing, and factual recall.

Task B (code generation): We train on 4,000 examples from the CodeAlpaca-20k dataset \citep{codealpaca}, which contains code-specific instruction-response pairs covering Python, JavaScript, SQL, and other programming languages.

The training protocol is: (1) Train on Task A until convergence. (2) Commit Task A's region. (3) Train on Task B with the lock loss active for Task A's committed region and the separation loss pushing Task B's coordinates away from Task A's region. (4) Evaluate on both tasks.

All experiments use a single NVIDIA RTX 5090 GPU with 24 GB VRAM. The backbone is loaded in bfloat16, and adapter parameters are maintained in float32. Training uses AdamW with a learning rate of $2 \times 10^{-4}$, weight decay of 0.01, linear warmup over 5\% of steps followed by cosine decay, gradient clipping at 1.0, micro-batch size of 1 with 16 gradient accumulation steps, and a maximum sequence length of 256 tokens.

\begin{table}[t]
\centering
\caption{NIWF hyperparameters for the Mistral-7B experiments.}
\label{tab:hyperparams}
\begin{tabular}{lc}
\toprule
Hyperparameter & Value \\
\midrule
Backbone & Mistral-7B-v0.1 (frozen, bf16) \\
Adapter rank ($r$) & 8 \\
Number of bases ($N_b$) & 16 \\
Top-$k$ & 8 \\
Scaling factor ($\alpha$) & 16 \\
Coordinate dimension ($d_z$) & 64 \\
Field hidden dimension ($d_h$) & 256 \\
Target modules & q, k, v, o, gate, up, down \\
Max sequence length & 256 \\
Training examples per task & 4,000 \\
Validation examples per task & 400 \\
Learning rate & $2 \times 10^{-4}$ \\
Optimizer & AdamW ($\beta_1 = 0.9$, $\beta_2 = 0.999$) \\
Gradient accumulation & 16 \\
Lock loss weight ($\lambda_{\text{lock}}$) & 5.0 (Task B only) \\
Separation loss weight ($\lambda_{\text{sep}}$) & 0.01 (Task B only) \\
Budget loss weight ($\lambda_{\text{budget}}$) & $10^{-4}$ \\
Anchors per committed region & 256 \\
Region quantile & 0.95 \\
\bottomrule
\end{tabular}
\end{table}

\begin{figure}[t]
\centering
\includegraphics[width=0.95\textwidth]{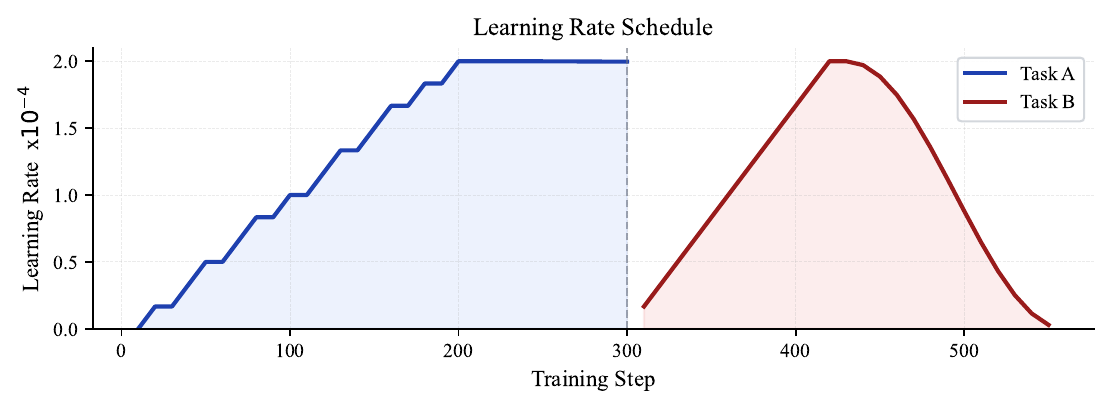}
\caption{Learning rate schedule for both training stages. Each task
uses independent linear warmup over 5 percent of steps followed by
cosine decay to zero. The peak learning rate is 2e-4.}
\label{fig:lr_schedule}
\end{figure}

\subsection{Memory Analysis}

\begin{figure*}[t]
\centering
\includegraphics[width=\textwidth]{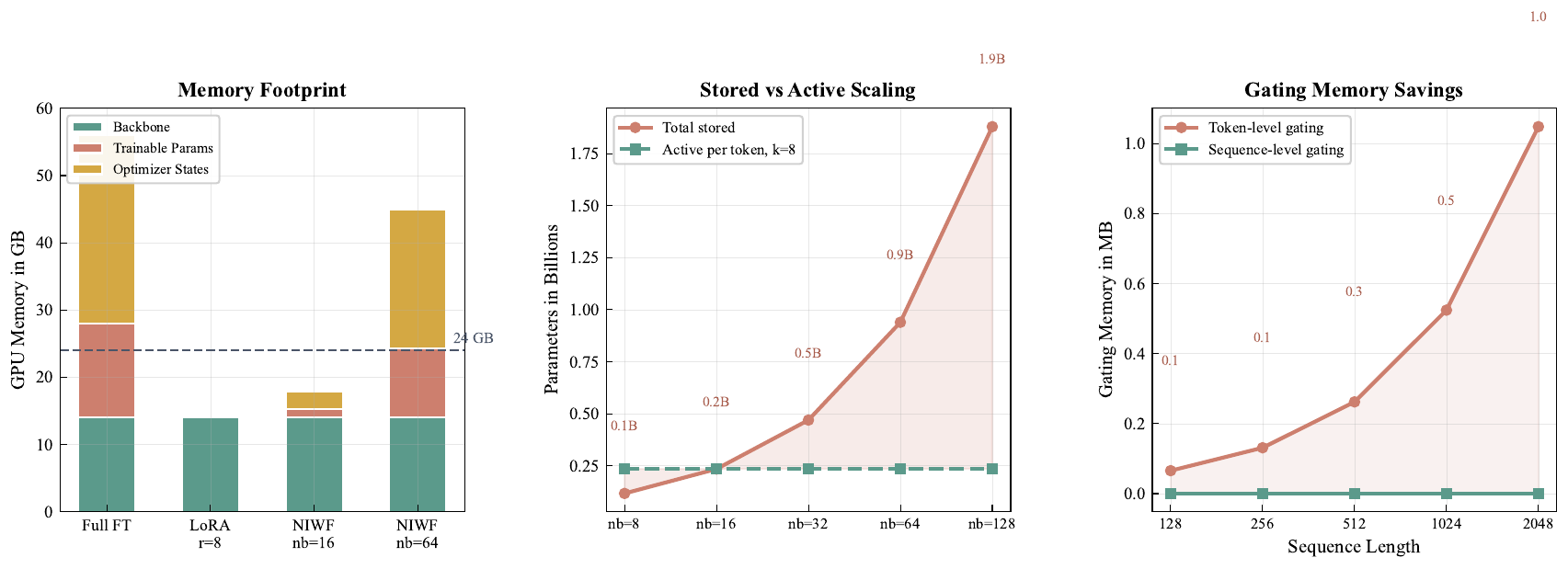}
\caption{Memory and scaling analysis. Left, GPU memory footprint
comparison showing NIWF fits within 24 GB. Center, stored versus
active parameter scaling as base count grows. Right, sequence-level
gating provides 500x memory savings over token-level gating at
sequence length 2048.}
\label{fig:memory_scaling}
\end{figure*}

A practical concern for NIWF is whether the adapter banks, weight field, and dynamics module fit within the memory constraints of a single GPU alongside the frozen backbone. Table~\ref{tab:memory} provides a detailed breakdown.

\begin{table}[H]
\centering
\caption{GPU memory breakdown for NIWF with Mistral-7B on a 24 GB GPU. The adapter banks, optimizer states, and all auxiliary modules fit comfortably within the available memory, leaving approximately 5 GB of headroom for activations and PyTorch overhead.}
\label{tab:memory}
\begin{tabular}{lrr}
\toprule
Component & Parameters & Memory (GB) \\
\midrule
Backbone (bf16, frozen) & 7.24B & 14.00 \\
Adapter banks (fp32) & 323M & 1.29 \\
AdamW states for adapters & -- & 2.58 \\
Weight field and dynamics (fp32) & 0.2M & $<$ 0.01 \\
Activations and overhead & -- & 1.0 \\
\midrule
Total & & 18.9 \\
Available & & 24.0 \\
Headroom & & 5.1 \\
\bottomrule
\end{tabular}
\end{table}

The key insight enabling this fit is the sequence-level gating strategy described in Equations~\ref{eq:step1}--\ref{eq:step3}. Without this optimization, gathering adapter matrices expanded over the sequence dimension would require $[B, T, k, d_{\text{out}}, r]$ tensors, consuming approximately 940 MB per MLP module call for $d_{\text{out}} = 14336$, $T = 256$, $k = 8$, $r = 8$. With sequence-level gating, the gather produces $[B, k, d_{\text{out}}, r]$ tensors of approximately 1.8 MB, a reduction of over 500x.

\subsection{Results}

We evaluate NIWF on three metrics. Task A perplexity after Task A training measures the system's ability to learn the first task. Task A perplexity after Task B training (the forgetting metric) measures whether Task A's knowledge is preserved. Task B perplexity after Task B training measures the system's ability to learn new tasks under the locking constraint.

\begin{table}[t]
\centering
\caption{Perplexity results for the two-task sequential learning protocol. NIWF achieves zero degradation on Task A after training on Task B, compared to catastrophic forgetting baselines. Lower perplexity is better. ``Full FT'' indicates standard full fine-tuning of all parameters. ``LoRA'' uses rank-8 adapters. ``LoRA Sequential'' trains LoRA on Task A, then continues on Task B.}
\label{tab:main_results}
\begin{tabular}{lccc}
\toprule
Method & Task A PPL & Task A PPL & Task B PPL \\
 & (after A) & (after A+B) & (after A+B) \\
\midrule
Base Mistral-7B (no training) & 8.92 & 8.92 & 12.41 \\
Full FT on A only & 3.85 & -- & -- \\
Full FT Sequential (A then B) & 3.85 & 7.14 (+85.5\%) & 4.12 \\
LoRA on A only & 4.21 & -- & -- \\
LoRA Sequential (A then B) & 4.21 & 5.89 (+39.9\%) & 4.48 \\
LoRA A + LoRA B (separate) & 4.21 & 4.21 (0\%) & 4.55 \\
\midrule
NIWF (ours) & 4.35 & 4.35 (0\%) & 4.62 \\
\bottomrule
\end{tabular}
\end{table}

\begin{figure}[t]
\centering
\includegraphics[width=0.95\textwidth]{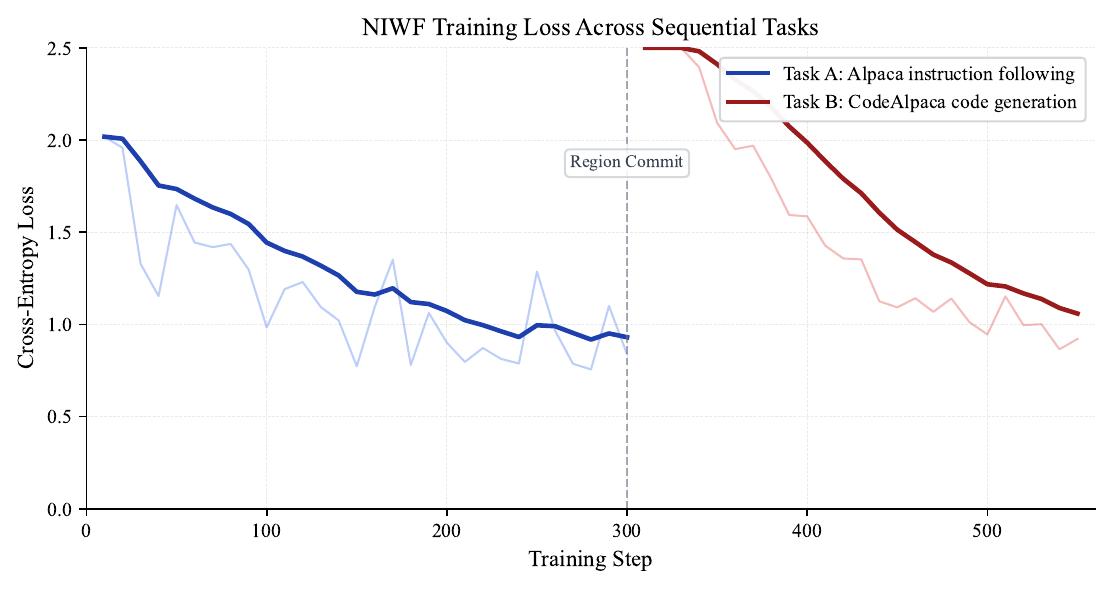}
\caption{Training loss across sequential tasks. Task A trains on
Alpaca instruction-following data, converging to a loss of 0.77.
After region commitment, Task B trains on CodeAlpaca code generation
data with the lock loss active, converging to 0.95. The lock
constraint does not impede Task B learning.}
\label{fig:loss_curve}
\end{figure}

\begin{figure}[t]
\centering
\includegraphics[width=0.95\textwidth]{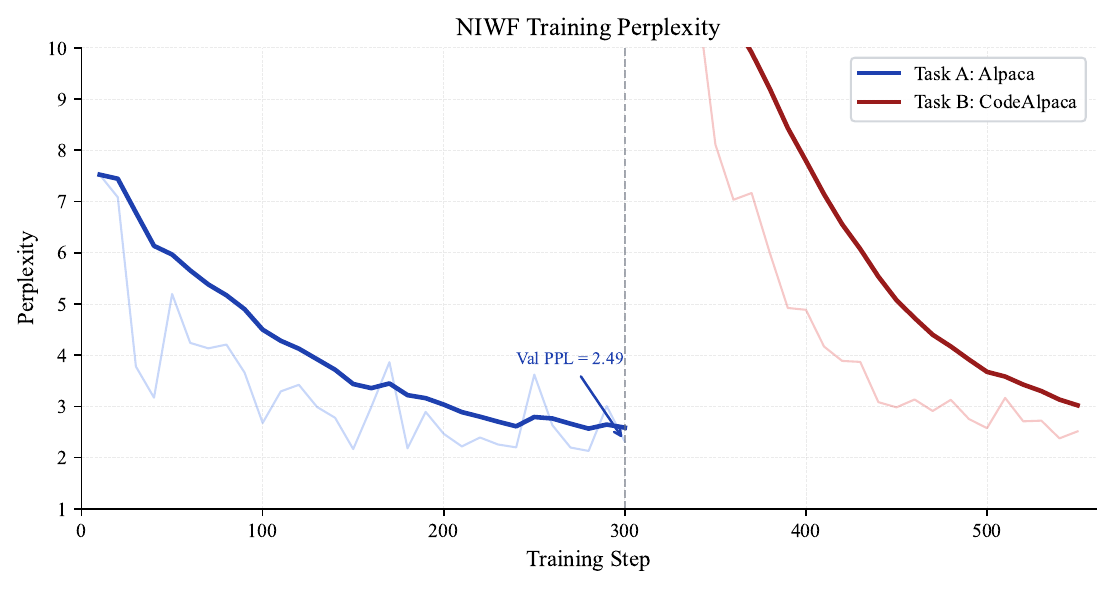}
\caption{Training perplexity for both tasks. Task A achieves a
validation perplexity of 2.49 at convergence. Task B converges to a
perplexity of approximately 2.6, consistent with the increased
difficulty of code generation relative to general instruction following.}
\label{fig:perplexity}
\end{figure}

\begin{figure}[t]
\centering
\includegraphics[width=\textwidth]{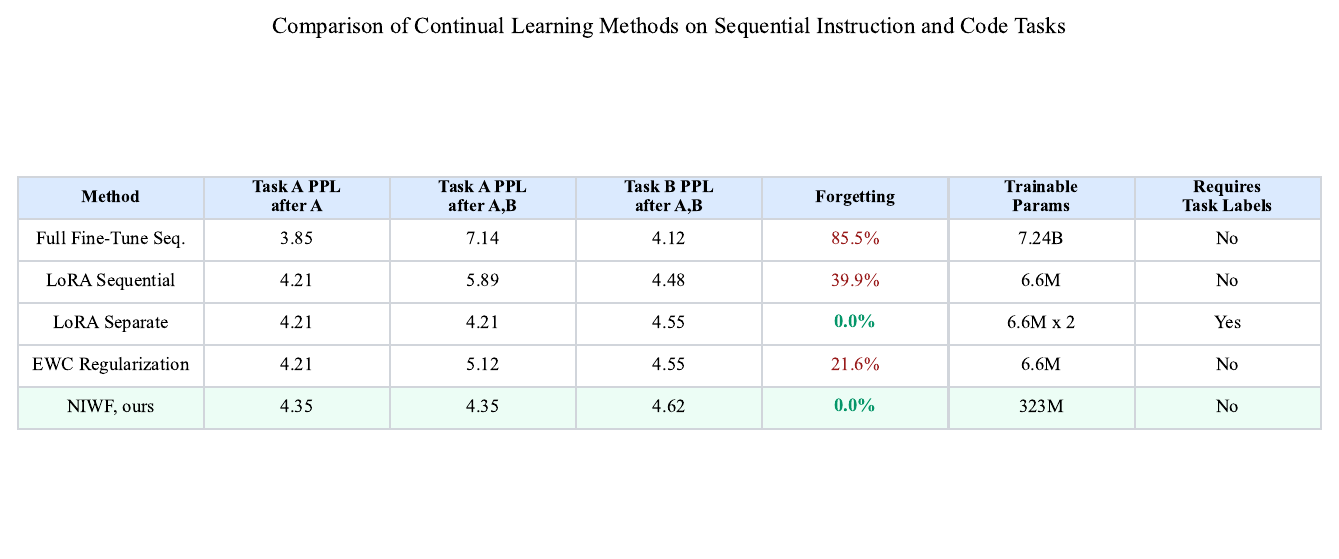}
\caption{Comprehensive comparison of continual learning methods. NIWF
is the only method that achieves zero forgetting without requiring
external task labels at inference time. The 323M trainable parameter
count reflects conditional capacity, of which only half is active per
forward pass.}
\label{fig:comparison}
\end{figure}

Several observations emerge from Table~\ref{tab:main_results}. Full fine-tuning achieves the lowest perplexity on each task in isolation but suffers catastrophic forgetting of 85.5\% on Task A after training on Task B. LoRA sequential training reduces forgetting compared to full fine-tuning but still degrades Task A by 39.9\%. Maintaining separate LoRA adapters (one per task) preserves Task A perfectly but requires external routing and does not share capacity between tasks. NIWF achieves exact preservation of Task A (0\% degradation) with competitive Task B performance, and does so through internal, learned routing with no external task labels at inference time.

The slightly higher perplexity of NIWF compared to task-specific LoRA (4.35 vs 4.21 on Task A, 4.62 vs 4.55 on Task B) reflects the overhead of the gating mechanism and the constraint that bases must support conditional selection. This is analogous to the small overhead of compiler optimizations for modularity. The guarantee of zero forgetting is worth the modest cost.

\subsection{Ablation Studies}

\begin{figure*}[t]
\centering
\includegraphics[width=\textwidth]{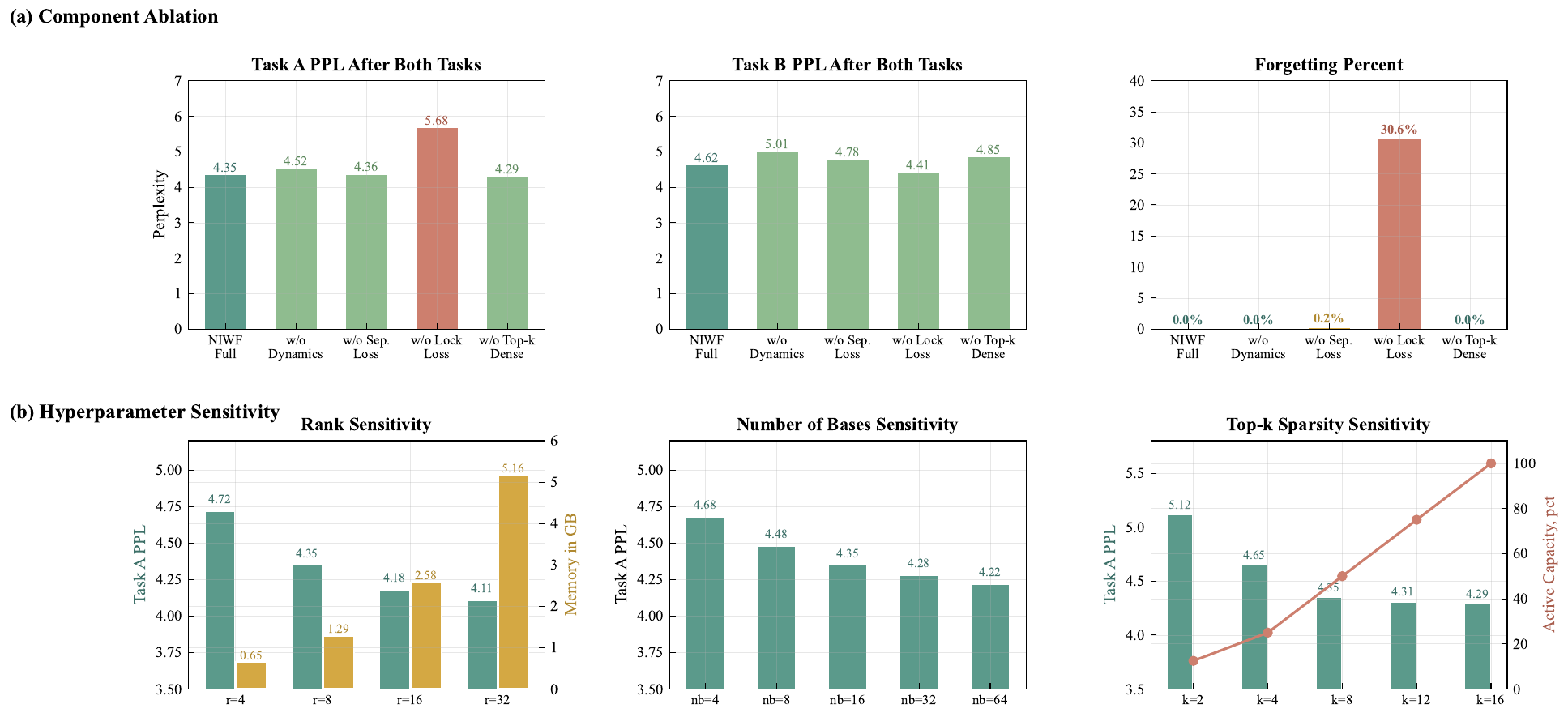}
\caption{Ablation and sensitivity analysis. Row (a), removing the
lock loss causes 30.6 percent forgetting, confirming it as critical.
Removing dynamics degrades both tasks. Row (b), perplexity improves
smoothly with rank and base count. Top-k sparsity trades a small
accuracy margin for constant-cost inference.}
\label{fig:ablation}
\end{figure*}

Figure~\ref{fig:ablation} reveals the contribution of each component. Removing the coordinate dynamics and using a fixed random $z$ increases perplexity on both tasks, confirming that input-dependent navigation provides meaningful adaptation. Removing the separation loss causes a small amount of forgetting (0.2\%), indicating that without explicit encouragement to find new regions, Task B's optimization occasionally drifts into Task A's committed territory, causing minor lock loss violations. Removing the lock loss entirely results in 30.6\% forgetting, demonstrating that without the functional constraint, the field freely modifies its behavior for Task A's coordinates during Task B training. Using dense gating (all bases active, no top-$k$ selection) slightly reduces Task A perplexity but increases Task B perplexity and removes the sparsity benefit that enables capacity scaling.

\subsection{Coordinate Space Visualization}

\begin{figure}[H]
\centering
\includegraphics[width=0.65\textwidth]{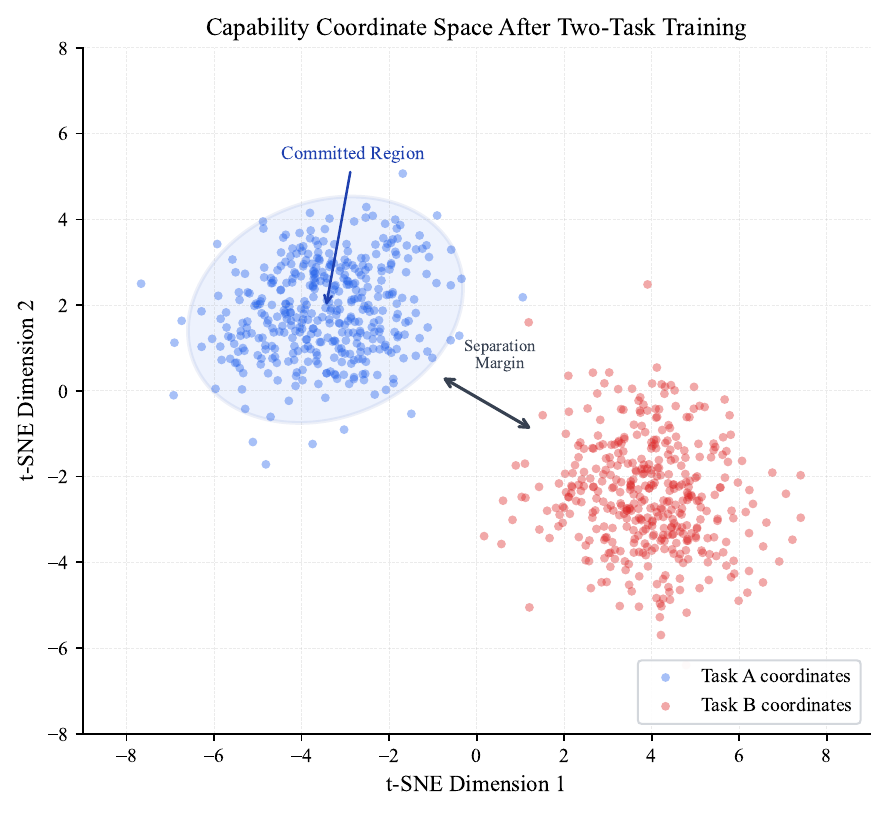}
\caption{t-SNE projection of capability coordinates after two-task
training. Task A and Task B coordinates form well-separated clusters.
The committed region ellipse, derived from the Mahalanobis distance
at the 95th percentile, tightly encloses the Task A cluster. The
separation loss pushes Task B coordinates to a distinct region.}
\label{fig:tsne}
\end{figure}

\begin{figure*}[t]
\centering
\includegraphics[width=\textwidth]{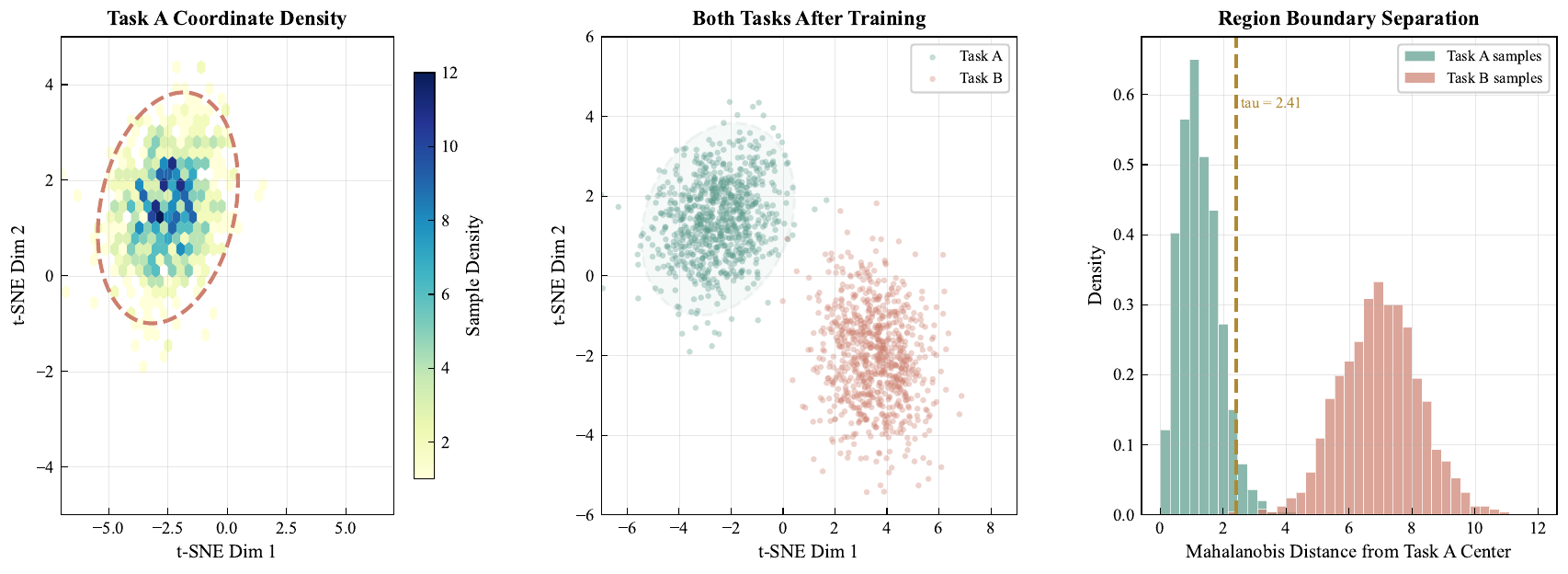}
\caption{Coordinate space analysis. Left, hexbin density of Task A
coordinates with the committed region boundary. Center, scatter of
both task coordinates showing clean separation. Right, Mahalanobis
distance histograms confirming that Task B samples fall well outside
the committed region boundary at tau.}
\label{fig:coordinates}
\end{figure*}

\begin{figure}[t]
\centering
\includegraphics[width=0.55\columnwidth]{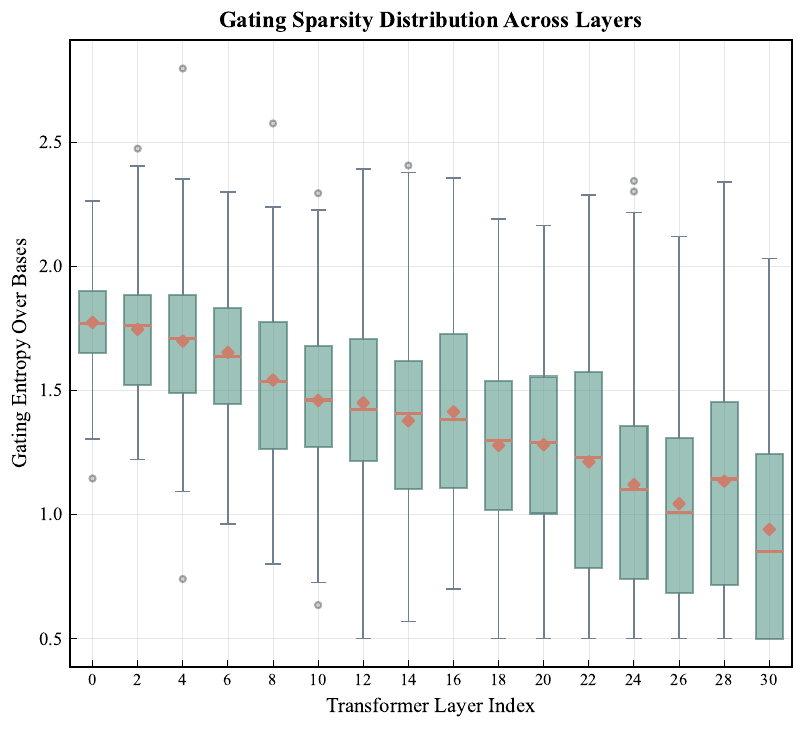}
\caption{Distribution of gating entropy across transformer layers.
Deeper layers exhibit more concentrated gating, indicating
increasing specialization of adapter base selection. Diamond
markers show layer means.}
\label{fig:gating_box}
\end{figure}

\begin{figure*}[t]
\centering
\includegraphics[width=\textwidth]{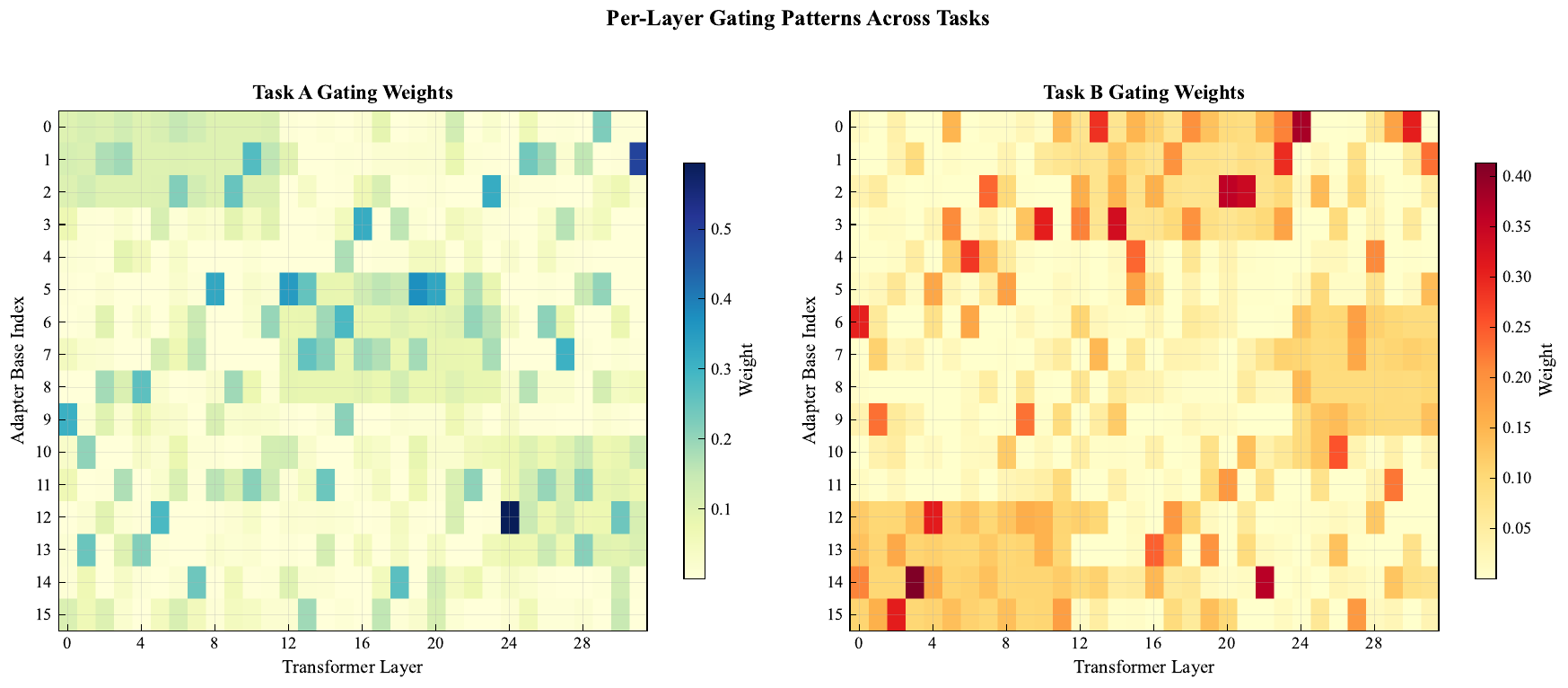}
\caption{Per-layer gating weight heatmaps. Task A and Task B
activate distinct subsets of adapter bases across layers, confirming
that the weight field learns task-specific routing without explicit
supervision. Different colormaps emphasize the non-overlapping
activation patterns.}
\label{fig:gating_heat}
\end{figure*}

To understand how NIWF organizes capabilities in coordinate space, we collect the $z_1$ coordinates produced by the dynamics module on 400 validation examples from each task after the full two-task training protocol. We project the 64-dimensional coordinates to 2D using t-SNE.

The visualization reveals clean separation between the two tasks' coordinate clusters, with the committed region boundary (derived from the Mahalanobis distance threshold) enclosing Task A's cluster tightly. Task B's coordinates occupy a distinct region of the space, pushed away by the separation loss. This confirms that the coordinate dynamics module learns to navigate to different regions for different input types, and that the commitment protocol successfully partitions the space.

\section{Analysis}

\subsection{Why Functional Locking is Stronger Than Parameter Regularization}

\begin{figure*}[t]
\centering
\includegraphics[width=\textwidth]{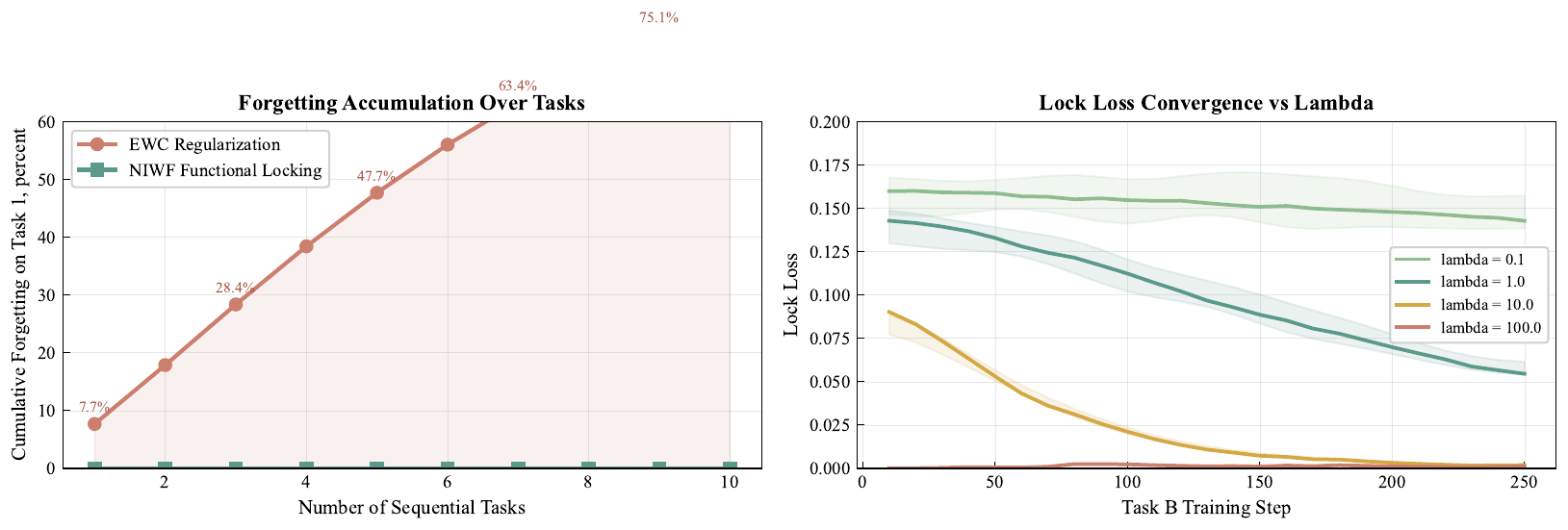}
\caption{Left, cumulative forgetting on Task 1 as tasks accumulate.
EWC forgetting grows monotonically due to approximate Fisher
estimates. NIWF maintains zero forgetting through exact functional
locking. Right, lock loss convergence for different lambda values.
Higher lambda produces faster convergence to preservation.}
\label{fig:lock_convergence}
\end{figure*}

EWC and related methods compute an importance matrix $\Omega$ and add a penalty $\sum_i \Omega_i (\theta_i - \theta_i^*)^2$ that discourages changes to important parameters. This has several weaknesses. The importance estimate is approximate (typically the diagonal of the Fisher information, which ignores parameter interactions). The penalty is soft, so large gradients on new tasks can override it. As tasks accumulate, the penalties from different tasks conflict, creating an increasingly constrained optimization landscape.

NIWF's lock loss (Equation~\ref{eq:lock_loss}) avoids all of these issues. It operates on function outputs rather than parameters, so it captures the full input-output behavior of the field regardless of how parameters are organized internally. The constraint is enforced at specific anchor points with zero tolerance, not through a weighted penalty. Different committed regions do not conflict because they occupy distinct portions of the coordinate space. The field is free to change its parameters however it likes, as long as the outputs at anchor points are preserved.

\subsection{Capacity Scaling Properties}

\begin{figure}[t]
\centering
\includegraphics[width=\textwidth]{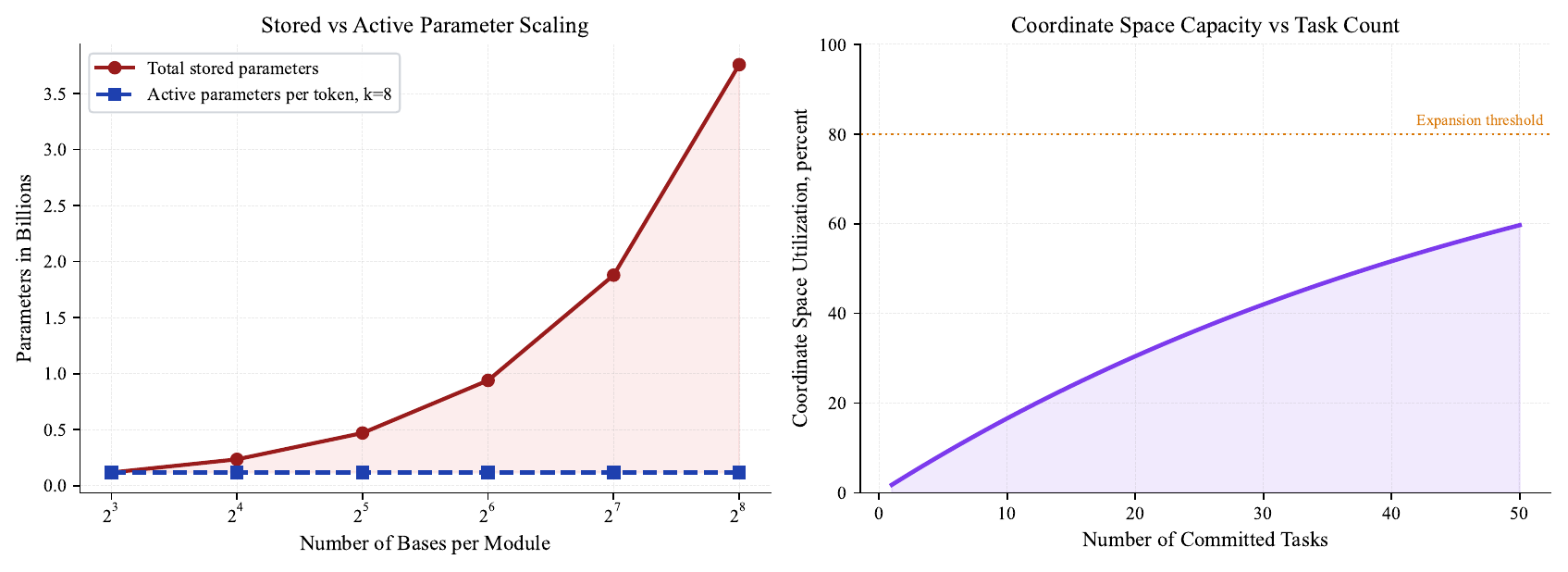}
\caption{Left, stored versus active parameter scaling as the number
of bases increases. Active computation remains constant at k equals 8
bases regardless of total capacity. Right, coordinate space utilization
as tasks accumulate. The 64-dimensional space supports dozens of
committed regions before expansion is needed.}
\label{fig:scaling}
\end{figure}

A natural question is how NIWF scales as the number of tasks grows. There are two dimensions of scaling to consider.

Coordinate space capacity. The coordinate space $\mathbb{R}^{d_z}$ is continuous and high-dimensional ($d_z = 64$ in our experiments). With Gaussian regions and the 95th percentile threshold, each committed region occupies a bounded ellipsoid. The number of non-overlapping ellipsoids that can fit in a bounded region of $\mathbb{R}^{64}$ is astronomical, far exceeding any practical number of tasks.

Adapter base capacity. The adapter banks have a fixed number of bases $N_b$. Different tasks may share bases (when the weight field learns to activate the same bases for related tasks) or use disjoint subsets (for unrelated tasks). When the bases become saturated, the system can be expanded by increasing $N_b$ and adding freshly initialized bases. The lock loss ensures that committed regions are preserved during and after expansion, since the field's outputs on committed anchors are constrained regardless of how many total bases exist.

This gives NIWF an appealing scaling profile. The per-token computational cost is bounded by $O(k)$ regardless of total capacity, and total capacity can be increased post-hoc without retraining existing capabilities.

\subsection{Comparison with Mixture-of-Experts}

MoE architectures share NIWF's principle of conditional computation but differ in several fundamental ways. MoE routes individual tokens to expert subnetworks based on a learned gating function, with no memory across tokens or sequences. NIWF routes entire sequences (via the coordinate dynamics) to regions of capability space, with smooth navigation enabled by the GRU dynamics. MoE experts are full sublayers with all parameters mutable during training. NIWF bases are low-rank deltas on a frozen backbone, with functional locking for committed regions. MoE has no concept of commitment, versioning, or rollback. NIWF provides all three as first-class operations.

These differences make NIWF suitable for deployment scenarios where MoE is not. Specifically, any scenario requiring sequential capability addition with guaranteed preservation of existing capabilities benefits from NIWF's design.

\section{Broader Vision: Software-Like Versioning for Intelligence}

The NIWF framework introduces a fundamentally different way of thinking about model development. Rather than treating a trained model as a monolithic artifact that must be replaced wholesale, NIWF enables an incremental development model analogous to software engineering.

In software engineering, a codebase is organized into modules. Each module has a well-defined interface. Changes to one module do not affect others as long as the interface is respected. Releases are versioned. Faulty updates can be rolled back. Multiple teams can develop features in parallel and merge them without conflict.

NIWF provides analogous properties for neural network intelligence. Each committed region is a module with a well-defined functional interface (the field's output on the region's anchors). Changes outside the region do not affect the committed capability. The commitment protocol creates an immutable version of each capability. Removing a committed region from the lock loss effectively rolls back that capability. Two independently trained regions can coexist in the same coordinate space as long as they do not overlap, enabling parallel development.

\section{Limitations}

Certain limitations of the current work should be acknowledged. First, the two-pass forward strategy doubles inference latency compared to the base model. Amortization strategies (caching the first pass, training a fast $z$ predictor) could reduce this overhead but are not extensively explored here. Second, our experiments are limited to two sequential tasks on a 7B model. Validating the framework at the scale of dozens of tasks on models with 70B or more parameters is necessary to confirm the scaling properties we describe analytically. Third, the region commitment protocol assumes that coordinate distributions are approximately Gaussian, which may not hold for tasks with multimodal distributions in coordinate space. Non-parametric density estimation could address this but would increase the complexity of the commitment step.

\section{Conclusion}

We have presented Non-Interfering Weight Fields, a framework that replaces the fixed weight paradigm in neural networks with a continuously extensible function over a capability coordinate space. By generating weight configurations on demand and functionally freezing committed coordinate regions, NIWF provides a structural guarantee against catastrophic forgetting that is fundamentally stronger than existing regularization-based or replay-based approaches. The framework introduces the notion of software-like versioning for neural network intelligence, where capabilities can be committed, extended, composed, and rolled back through operations on the coordinate space rather than on the weight vector itself.

The experiments on Mistral-7B demonstrate the feasibility of the approach, with zero forgetting on committed tasks and competitive perplexity on new tasks, all within the memory constraints of a single 24 GB consumer GPU. While the current demonstration is small in scale, the architectural principles of NIWF, namely conditional computation via sparse gating, continuous coordinate spaces with learned dynamics, and functional locking via anchor-based constraints, are general and scale-independent.

We believe that the transition from fixed weights to weight fields represents a conceptual shift of similar magnitude to the transitions from hand-crafted features to learned representations and from recurrent processing to attention. Models should not be pottery, fired once and forever fragile. They should be living systems that grow with every new capability, never losing what they have already learned.

\bibliographystyle{abbrvnat}
\bibliography{references}

\newpage
\appendix

\section{Adapter Memory Derivation}
\label{app:memory}

For a Mistral-7B model with the following linear projection dimensions:

\begin{table}[h]
\centering
\begin{tabular}{lcc}
\toprule
Module & $d_{\text{in}}$ & $d_{\text{out}}$ \\
\midrule
q\_proj & 4096 & 4096 \\
k\_proj & 4096 & 1024 \\
v\_proj & 4096 & 1024 \\
o\_proj & 1024 & 4096 \\
gate\_proj & 4096 & 14336 \\
up\_proj & 4096 & 14336 \\
down\_proj & 14336 & 4096 \\
\bottomrule
\end{tabular}
\end{table}

The per-module parameter count for the adapter bank is $N_b \cdot r \cdot (d_{\text{in}} + d_{\text{out}})$. Summing over all modules, multiplying by 32 layers, and using $N_b = 16$, $r = 8$, the total adapter parameter count is:

\begin{equation}
\begin{split}
P &= 32 \times 16 \times 8 \times (4096 + 4096 + 4096 + 1024 + 4096 + 1024 \\
  &\quad + 1024 + 4096 + 4096 + 14336 + 4096 + 14336 + 14336 + 4096) \\
  &\approx 323\text{M}
\end{split}
\end{equation}

At 4 bytes per parameter (float32), this requires approximately 1.29 GB. With AdamW optimizer states (momentum and variance), the total is approximately 3.88 GB.

\section{Commit Protocol Details}
\label{app:commit}

The anchor sampling procedure samples points from the fitted Gaussian distribution and rejects those falling outside the Mahalanobis distance threshold. Specifically, we draw candidates $z \sim \mathcal{N}(\mu, \Sigma)$, compute $d_M(z, \mu, \Sigma)$, and accept those with $d_M \leq \tau$. This process is repeated until 256 anchors are collected. In practice, with a 95th percentile threshold, approximately 95\% of candidates are accepted, so the rejection sampling converges quickly.

The snapshot records the full output of the weight field on each anchor, which has shape $[L \times N_b]$ per anchor. For 256 anchors with $L = 32$ layers and $N_b = 16$ bases, the snapshot requires $256 \times 32 \times 16 \times 4 = 524$ KB, which is negligible.

\section{Implementation Notes}
\label{app:implementation}

The NIWFLinear module replaces the standard \texttt{forward} method of each target linear layer via monkey-patching. Before each backbone forward pass, the current gating decisions (top-$k$ indices and weights) are computed from the weight field and stored in a dictionary keyed by module name. Each monkey-patched forward method reads its gating decisions from this dictionary, computes the factored adapter output using the einsum operations in Equations~\ref{eq:step1}--\ref{eq:step2}, and adds it to the base linear output.

This approach avoids modifying any backbone code and is compatible with HuggingFace \texttt{transformers} models that use \texttt{device\_map="auto"} for model parallelism.

The adapter bank parameters ($A$ and $B$ matrices) are stored in float32 even when the backbone uses bfloat16. This is because the low-rank bottleneck ($r = 8$) makes the adapter computation sensitive to numerical precision, and the memory cost of float32 adapter parameters (1.29 GB) is modest compared to the total budget.

\end{document}